\documentclass[12pt]{article}

\usepackage[T2A]{fontenc}
\usepackage[cp866]{inputenc}  

\usepackage{newlfont}
\usepackage{supertabular}
\usepackage{amsmath}
\usepackage{amssymb}
\usepackage{amsfonts}
\usepackage{amsthm}
\usepackage{bm}
\usepackage{mathtools}
\usepackage{pb-diagram}
\usepackage{righttag}
\usepackage{eqnarray}
\usepackage{subeqnarray}
\usepackage{algorithmic}
\usepackage{epsfig}
\usepackage{subfigure} 
\usepackage{emlines2}
\usepackage{rotating}
\usepackage{multirow}
\usepackage{graphicx}
\usepackage{float}
\usepackage{makecell}

\usepackage[small,sc]{caption2} 
\usepackage{cite} 
\usepackage[english, russianb]{babel}

\usepackage{color}

\usepackage{booktabs}
\usepackage{url}
\usepackage{float}

\usepackage{fancybox}

\theoremstyle{plain}

\theoremstyle{definition}
\newtheorem{example}{Пример}

\newtheoremstyle{break}
  {9pt}
  {9pt}
  {\itshape}
  {}
  {\bfseries}
  {.}
  {\newline}
  {}

\theoremstyle{break}

\makeatletter

\renewcommand{\section}{\@startsection{section}{1}%
             {\parindent}{3.5ex plus 1ex minus .2ex}%
             {2.3ex plus.2ex}{\normalsize\bf}}

\renewcommand{\subsection}{\@startsection{subsection}{2}%
             {\parindent}{3.5ex plus 1ex minus .2ex}%
             {2.3ex plus.2ex}{\normalsize\bf}}

\renewcommand{\subsubsection}{\@startsection{subsubsection}{3}%
             {\parindent}{3.5ex plus 1ex minus .2ex}%
             {2.3ex plus.2ex}{\normalsize\bf}}

\renewcommand{\paragraph}{\@startsection{paragraph}{4}%
             {\parindent}{3.5ex plus 1ex minus .2ex}%
             {2.3ex plus.2ex}{\normalsize\bf}}

\renewcommand{\@biblabel}[1]{#1.}

\makeatother

\graphicspath{{Picts/}}

\newcounter{partnumber}

\renewcommand{\tablename}{Table}
\renewcommand{\figurename}{Fig.}

\renewcommand{\refname}{СПИСОК ИСПОЛЬЗОВАННЫХ ИСТОЧНИКОВ}

\makeatletter
\@addtoreset{equation}{section}
\renewcommand{\theequation}{\arabic{equation}}

\@addtoreset{table}{section}

\makeatother

\newcounter{fragm}
\newcounter{subfragm}[fragm]

\newcounter{myremark}[section]

\newcounter{myalgorithm}[section]

\begin{document}


\begin{center}
\textbf{uSF: Learning Neural Semantic Field with Uncertainty}
\end{center}

\begin{center}
V.S. Skorokhodov$^{1}$, D.M. Drozdova$^{1}$, D.A. Yudin$^{1,2,3*}$ \\
$^{1}$Moscow Institute of Physics and Technology, Moscow, Russia \\
$^{2}$AIRI (Artificial Intelligence Research Institute), Moscow, Russia \\
$^{3}$Federal Research Center ``Computer Science and Control'', Moscow, Russia \\
$^{*}$e-mail: 
yudin.da@mipt.ru\\
\end{center}

\bigskip

\small
\textbf{Abstract.}
Recently, there has been an increased interest in NeRF methods which reconstruct differentiable representation of three-dimensional scenes. One of the main limitations of such methods is their inability to assess the confidence of the model in its predictions. In this paper, we propose a new neural network model for the formation of extended vector representations, called uSF, which allows the model to predict not only color and semantic label of each point, but also estimate the corresponding values of uncertainty. We show that with a small number of images available for training, a model that quantifies uncertainty performs better than a model without such functionality. Code of the uSF approach is publicly available at \url{https://github.com/sevashasla/usf/}.


\section{Introduction}

Over the past few years, many researchers have focused on the development of the field of differentiable scene representation. This development began with the appearance of the first NeRF method \cite{original_nerf}, which demonstrated the photo-realistic quality of novel views generation. Various modifications of the original method allow us to additionally solve such problems as depth estimation \cite{nerfingmvs}, semantic segmentation\cite{semantic_nerf,semantic_ray}, object detection \cite{nerf_rpn}. Among the disadvantages of the NeRF methods one can note high computational complexity \cite{instant_ngp,f2_nerf}, the need for a large number of images \cite{active_nerf,depth_nerf} in the training dataset and the inability to estimate the confidence of the model in its predictions. Quantifying uncertainty is a crucial aspect for minimizing possible risks for tasks in various fields, for example robotics \cite{nerf_applications} and autonomous driving \cite{uncertainty_autonomus_driving, sellat2022advanced}.

\begin{figure}[t!]
\centering
\includegraphics[width=\linewidth]{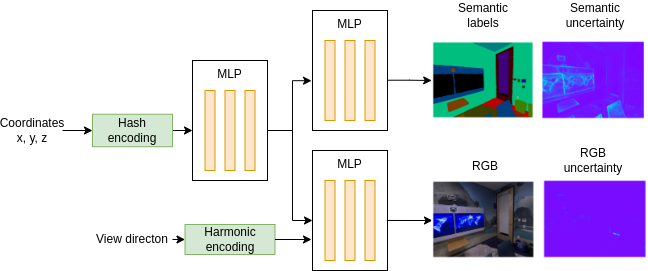}
\caption{We propose the model named uSF to predict both color and semantic labels and estimate corresponding uncertainty.}
\label{fig:scheme_usf}
\end{figure}

As shown in Fig. \ref{fig:scheme_usf} we propose a novel NeRF-based method that reconstructs the neural radiance and the neural semantic fields. The previously proposed NeRF methods only considered estimating the uncertainty for color predictions, i.e. regression task \cite{nerf_wild,stochastic_nerf,conditional_flow_nerf}. Our method additionally introduces the possibility of quantifying uncertainty for predicted semantic labels, i.e. classification task. There are two types of uncertainty: aleatoric and epistemic. Aleatoric uncertainty reflects the noise inherent in the input data. Epistemic uncertainty reflects what the model does not know due to a lack of data, too simple architecture, or inefficient training process. In both cases, aleatoric uncertainty is estimated. The high value of uncertainty may also indicate errors occurred in the data preparation process, for example, poor quality of labelling in segmentation tasks. We choose aleatoric uncertainty over epistemic uncertainty because quantifying aleatoric uncertainty does not require major changes in network architecture and multiple inferences or training. Adding uncertainty estimation to the algorithm pipeline does not degrade the quality of reconstruction with a sufficient number of images and improves the quality with a small training dataset. Our method is also computationally efficient because it uses trainable positional encoding based on hashing \cite{instant_ngp}. We summarize our main contributions as follows:

\begin{itemize}
    \item{
        We have proposed a new neural network model called uSF, which allows to add the ability to quantify the uncertainty for predicted neural semantic field, while maintaining quantifying uncertainty for neural radiance field.
    }
        \item{
         Due to the use of trainable positional encoding based on hashing, our approach is able to noticeably reduce the running time of the algorithm.
    }
    \item{
        We have shown that quantifying uncertainty for both color and semantic labels predictions improves the quality of semantic 3D-scene reconstruction with a small number of images in the training set from open Replica Dataset.
    }
\end{itemize}

\renewcommand{\figurename}{Fig.}

\section{Related Work}

\subsection{Neural Radiance Fields}

Due to the great popularity of the NeRF method, the researchers have developed a large number of works expanding the possibilities of the original method. Some papers address the inefficiency of training and rendering \cite{instant_ngp,f2_nerf,mimo_nerf,steer_nerf}. The other methods make it possible to apply the NeRF under certain restrictions: limited number of images in the training dataset \cite{active_nerf,depth_nerf,pixel_nerf,mvsnerf}, lack of information about camera poses \cite{nope_nerf}, unlimited scene space \cite{merf}, dynamic scenes \cite{nerf_ds,d_nerf}.

\subsection{Neural Semantic Fields}

There are NeRF methods that can predict not only the color at each point of the scene, but also corresponding semantic labels. The Semantic-NeRF \cite{semantic_nerf} method reconstructs the neural semantic field with the help of an additional head in the model architecture. The process of getting semantic labels at a point is similar to getting a color. This method is highly robust to the presence of outliers in the training data, yielding quality results even with 90\% noise.

Another approach to segmentation is shown in the NeRF-SOS \cite{fan2022nerf} method. The work successfully combines self-supervised 2D visual features for 3D semantic segmentation. Separate head outputs for segmentation are trained similar to Dino-ViT \cite{caron2021emerging}. NeSF \cite{nesf} is the method that produces generalizable neural semantic fields. The authors propose to firstly recover 3D density fields upon which 3D semantic segmentation model is trained supervised by posed 2D semantic maps. The pre-trained NeRF model is used to sample the volumetric density grid for further conversion into a semantic-feature grid. Semantic Ray \cite{semantic_ray} also has generalization ability and uses Cross-Reprojection Attention module to obtain such result.

\subsection{Uncertainty Estimation}

To estimate the aleatoric uncertainty, a probability distribution is imposed on the outputs of the model. This approach assumes that there is noise in the input data, the distribution parameters of which can be predicted as some function of the data. To predict epistemic uncertainty, it is a common practice to impose a distribution on the model parameters (Bayesian neural networks (BNN)~\cite{bnn,bnn_book,dropout_inference}). The disadvantage of these methods for quantifying epistemic uncertainty is the requirement to train the model several times and make multiple inferences. There are also approaches that simultaneously estimate both types of uncertainties~\cite{evidential_regres,evidential_class,kendall2017uncertainties}.

Uncertainty quantification in various implementations has already been applied in several NeRF models. The Active NeRF ~\cite{active_nerf} is designed to deal with the limited number of images in the training set. In this method, the estimated aleatoric uncertainty is used for adding new frames to the training data during the iterations of active learning. In the NeRF-W ~\cite{nerf_wild} method aleatoric uncertainty is used to identify non-static objects that should be removed from the prediction. This approach allows you to obtain the scene reconstruction consisting only of static objects. The two following methods focus on quantifying of epistemic uncertainty. In the S-NeRF ~\cite{stochastic_nerf} method, it is assumed that all possible neural radiance fields for the scene are distributed according to a prior distribution, on the basis of which and the input data the model predicts a posterior distribution, with which the uncertainty is estimated. The CF-NeRF\cite{conditional_flow_nerf} method works similarly, but does not impose any preliminary restrictions on the distribution. Recently in Bayes'
Rays work \cite{bayes_rays_uncert} the authors have proposed a new method to estimate epistemic uncertainty that does not require any changes during training process. All of these methods try to quantify uncertainty for color or depth predictions, while in this paper we propose the approach for estimating corresponding uncertainty for semantic label predictions.

\section{Methodology}

\begin{figure}[t!]
    \includegraphics[width=\linewidth]{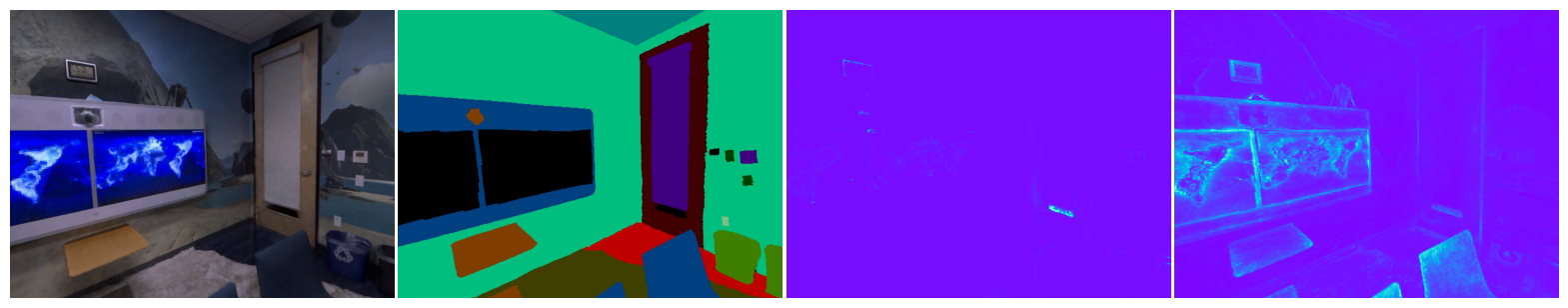}
    \includegraphics[width=\linewidth]{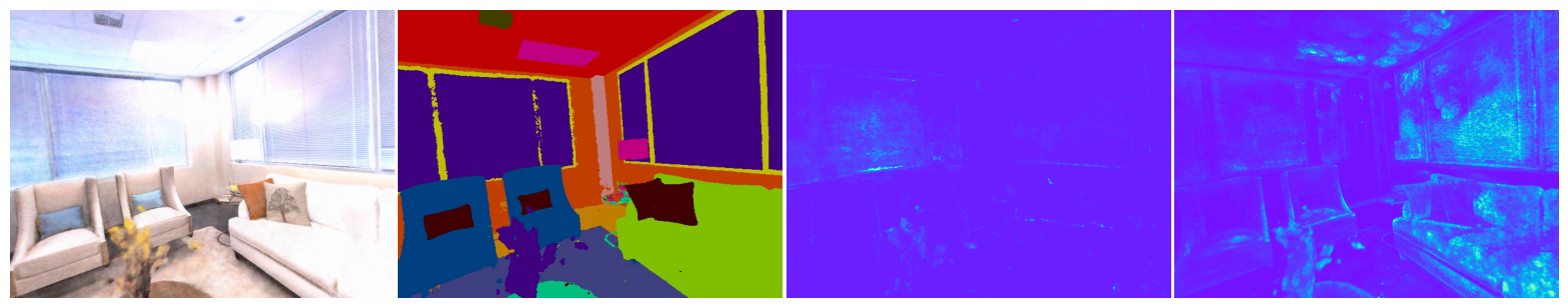}
    \caption{From left to right we show the predicted color, predicted semantic labels, rgb uncertainty and semantic uncertainty during training process.}
    \label{preds}
\end{figure}

\subsection{Neural Network Model}

\renewcommand{\figurename}{Fig.}
\begin{figure}[t!]
    \begin{center}
        \includegraphics[width=0.9\linewidth]{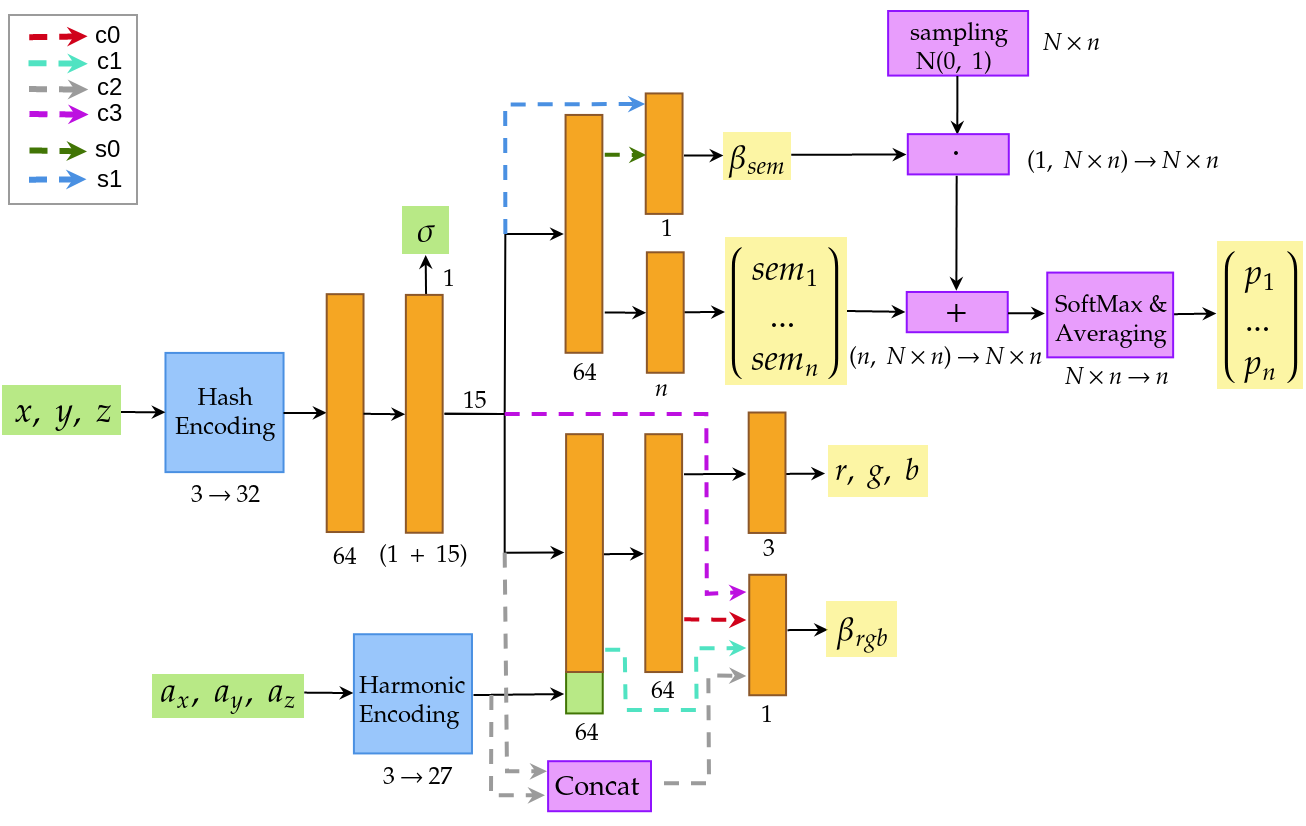}
        \caption{uSF architecture. Orange rectangles are Linear + ReLU layers. Yellow rectangles are outputs of our model. We study different variants of our architecture in terms of the color (c0-c3) and the semantic (s0-s1) branches.}
        \label{fig:network}
    \end{center}
    \medskip 
    \footnotesize   
\end{figure}

The architecture of our model is MLP with ReLU activation functions and a hidden state dimension equal to 64 as shown in the Fig. \ref{fig:network}. Similar to Active NeRF \cite{active_nerf} and NeRF-W \cite{nerf_wild} pipeline we add an additional head to predict the color variance $\beta_{rgb}$ which we use as rgb uncertainty. In our work we assume that it depends on the view direction.

We add two heads for semantic segmentation task. One outputs mean values $(sem_1, \dots, sem_n)$, and the other predicts the variance $\beta_{sem}^2$. We consider outputs of semantic heads to depend only on coordinates of the point, but not on view direction.

For semantic head we make MLP smaller than for color head with only two layers. We also use ReLU activation function and set the dimensionality of the hidden state to 64. 

\subsection{Uncertainty for Neural Radiance Field}

As described in the previous section the color part of the architecture consists of two heads \cite{kendall2017uncertainties}. One outputs the mean value parameter of some normal distribution which we consider as the color at the point. The other head outputs the variance which depends both on the coordinates of the point and on the view direction corresponding to it. To get the final value of uncertainty, the softplus activation function is used:

\begin{equation}
    {\bar\beta}^2 = {\beta_0}^2 + \log(1 + \exp(\beta^2)).
\end{equation}

The $\beta_0$ parameter is the minimum possible value of rgb uncertainty. To obtain the uncertainty values at each point of 3D space of the scene we use volume render procedure. The example of the predicted rgb uncertainty is shown in the Fig.~\ref{preds}

\subsection{Uncertainty for Neural Semantic Field}

From the semantic head our model outputs the mean values ($sem_i$) and the variance ($\beta_{sem}$) of semantic logits. We assume that $logit_i \sim \mathcal{N}(sem_i, \beta_{sem}^2)$. We want to use a CrossEntopy for the network optimization, but there are some difficulties of obtaining the exact class probabilities because one can't calculate such mathematical expectation~\ref{eq: mathexp}.
\begin{equation}\label{eq: mathexp}
    \widehat{p_i} = \mathbb{E}\,\text{SoftMax}(\{logit_1, \dots, logit_n\})_i
\end{equation}

In order to deal with such problem we use an approximation. We generate $N = 10$ sets of logits from the corresponding normal distributions, then count SoftMax separately for each set, and then average the obtained results~\ref{eq: approx_mathexp}:
\begin{equation}\label{eq: approx_mathexp}
    p_{pred, i} \approx \frac{1}{N}\widehat{p_i}
\end{equation}

During calculation we use the "reparametrization trick". Let $\xi \sim \mathcal{N}(a, \sigma^2)$, and $\eta \sim \mathcal{N}(0, 1)$. Then $\xi~{\buildrel d \over =}~a + \sigma \cdot \eta$. Then to generate logits, we firstly generate random values from $\mathcal{N}(0, 1)$, and then convert them to the form we want by multiplying by the standard deviation ($\beta_{sem}$) and adding the mean ($sem_{i}$). The example of the predicted semantic uncertainty is shown in the Fig. \ref{preds}

\subsection{Learning Approach}

To train our model we use the following loss function~\ref{eq: loss}, which consists of three terms:

\begin{equation}\label{eq: loss}
    Loss = \omega\,L_{rgb} + \lambda\,L_{semantic} + (1 - \omega)\,L_{uncert},
\end{equation}

where:

\begin{equation}
    L_{rgb} = \frac{1}{N}\sum_{i=1}^N\left(C(r_i) - rgb_i\right)^2,
\end{equation}

where $C(r_i)$ is predicted color for ray $r_i$, $rgb_i$ is ground truth color for ray $r_i$.

\begin{equation}
    L_{semantic} = CrossEntropy(p_{pred}, semantic_{gt}),
\end{equation}

where $p_{pred}$ is a set of predicted probabilities of semantic classes, $semantic_{gt}$ is a set of ground truth semantic labels.

\begin{equation}
    L_{uncert} = \frac{1}{N}\sum_{i=1}^N\bigg{(}\frac{||C(r_i) - rgb_i||^2}{2\beta_{rgb}(r_i)^2} + 
    \frac{1}{2}\log{\beta_{rgb}(r_i)^2} + \frac{\eta}{N_{si}}\sum_{j=1}^{N_{si}}\alpha(r_i(t_j))\bigg{)},
\end{equation}

where $C(r_i)$ is predicted color for ray $r_i$, $rgb_i$ is ground truth color for ray $r_i$, $\beta_{rgb}(r_i)^2$ is predicted variance for color for ray $r_i$, $N_{si}$ is number of points $t_j$ sampled on ray $r_j$, $\alpha(r_i(t_j))$ is point weight from volume rendering procedure ~\cite{original_nerf} for ray $r_i$ in point $t_j$, $\eta$ is a hyperparameter which controls the strength of regularization.

Parameters $\omega, \lambda$ are hyperparameters. We have decided to make the different terms $L_{rgb}$ and $L_{uncert}$, because they are both responsible for learning to make color predictions. In our experiments we change only $\omega$ and $\lambda$, and set $\alpha = 10^{-3}$.

We use Adam optimizer with $lr = 0.003, \beta=(0.9, 0.99), eps=10^{-15}$. During training the learning rate ($lr$) is changed according to equation~\ref{eq:lr}. This equation is based on the fact that at the beginning learning is unstable due to big modulus of the gradient. Also at the end learning rate should be smaller to obtain better results.

\begin{equation}\label{eq:lr}
    lr(i) = lr(0) \cdot 0.1^{\min\left(1, \frac{i}{N_i}\right)} \cdot \min \left(1, 10^{-3} + \frac{i}{N_{wi}} \right),
\end{equation}
where $i$ is iteration index, $N_i$ is total number of iterations, $N_{wi}$ is number of warmup iterations.

We use an early stopping during training. We firstly calculate mean of last 3 LPIPS values on eval dataset and then if $|\text{current LPIPS} - \text{mean}(\text{last 3 LPIPS})| < 10^{-3}$ is two times in a row, the training process is stopped.

We use PSNR, SSIM, LPIPS as the main metrics for the quality of color prediction like in other popular articles on 3D reconstruction. We use mIoU and accuracy as the main metric to qualify semantics maps prediction.

\section{Experiments}

\subsection{Hardware and Software Setup}

We have performed experiments on NVIDIA GeForce 2080 Ti, NVIDIA GeForce 2080, NVIDIA GeForce 3060 graphics cards. The model takes about 6GB of video memory during training. According to our estimates, full training lasts approximately 4-5 hours for one scene. 

\subsection{Datasets and Environments}

As the main dataset, we use Replica \cite{replica19arxiv}, which is a set of synthetic room and office scenes with corresponding semantic masks. It is the photorealistic indoor dataset \cite{yudin2022hpointloc}, which is widely used for scene reconstruction and mapping tasks, intelligent agent navigation, etc.

We use a pre-processed dataset from the Semantic NeRF \cite{semantic_nerf} method with image size of $640 \times 480$. Each scene contains 900 images with different view directions. We have performed all our experiments on scenes "room\_0", "room\_1" and "office\_0" since they represent a typical home and office environment. For the dataset we have 2 different splitting options:
\begin{itemize}
    \item{
        Small: 27 images for training, 112 - for validation, 225 - for test, 
    }
    \item{
        Large: 562 images for training, 112 - for validation, 225 - for test.
    }
\end{itemize}

\subsection{Architecture Choice}

There are several options of the architectures for our method which are shown in Fig.~\ref{fig:network}. The options differ in the different number of layers and the presence of dependence on the view direction. To properly select the optimal hyperparameters $\lambda$ and $\omega$, we firstly have manually found rough estimates for the optimum, and then used 10 runs of Bayesian hyperparameter fitting for each architecture. We have chosen the search interval for optimal hyperparameters: $\sim Uniform[10^{-5}, 10^{-3}]$ for $\omega$ and  $\sim Uniform[10^{-3}, 10^{-1}]$ for $\lambda$.

During the search we use only one scene room\_0 from Replica Dataset, because, as practice shows, the quality of reconstruction on it is slightly worse than on the rest of the scenes. We use only small (27) training set for all experiments.

\renewcommand{\tablename}{Tab.}
\begin{table}[t]
    \centering
    \begin{tabular}{ cccc } \hline 
    Network architecture & best $\lambda$ & best $\omega$ & mIoU \\ \hline
    uSF (c0, s0) & {\textbf{7.25$\cdot10^{-2}$}} & {\textbf{1.49}$\cdot 10^{-5}$} & {\textbf{0.570}} \\ 
    uSF (c1, s0) & {8.47$\cdot10^{-2}$} & {5.51$\cdot 10^{-5}$} & {0.536} \\ 
    uSF (c2, s0) & {6.01$\cdot10^{-2}$} & {28.95$\cdot 10^{-5}$} & {0.5272} \\ 
    uSF (c3, s0) & {7.89$\cdot10^{-2}$} & {2.13$\cdot 10^{-5}$} & {0.5102} \\ 
    uSF (c0, s1) & {5.67$\cdot10^{-2}$} & {1.45$\cdot 10^{-5}$} & {0.5579} \\ \hline
    \end{tabular}
    \caption{Hyperparameter values at which the best quality of reconstruction is achieved on a given architecture.}
    \label{tab:architecture_choice}
\end{table}

As shown in Tab. \ref{tab:architecture_choice} the best results are achieved with uSF (c0, s0). The experiments have also proved, that is it better to consider uncertainty $\beta_{rgb}$ as dependent on the view direction.

\subsection{Positional Encoding Choice}

We aim to create an effective implementation of our method without deterioration in quality. To achieve this, we have used the hash positional encoding which is described in Instant-NGP \cite{instant_ngp} method. It allows us to significantly reduce the number of layers in the architecture and the size of dimensions. Hidden layer dimension is 64 instead of 256, number of dense layers is 5 instead of 8.

Additionally, we have performed experiments to compare the networks with different type of positional encoding and architectures. The model with efficient architecture using hash positional encoding has shown better results in terms of speed and quality than the model from the original NeRF \cite{original_nerf} (8 dense layers, 1 skip-connection, hidden dimension is 256) using harmonic (frequency) positional encoding. The overall comparison results can be seen in Table \ref{tab:table_encoding}.

\renewcommand{\tablename}{Tab.}
\begin{table}[t]
    \centering
    \begin{tabular}{ c c c c c c c c c } \hline
    Scene & mIoU & PSNR & SSIM & LPIPS & Accuracy & $\text{T}_\text{render}$, s & Encoding & Architecture \\ \hline
    \multirow{4}{*}{room\_0} & \textbf{0.961} & 31.024 & 0.902 & 0.109 & \textbf{0.980} & 1.993 & hash & \multirow{4}{*}{Small} \\ 
     & 0.749 & 22.538 & 0.667 & 0.562 & 0.823 & \textbf{0.893} & freq & \\ 
     & - & \textbf{31.067} & \textbf{0.905} & \textbf{0.104}  & - &1.073 & hash & \\ 
     & - & 22.486 & 0.666 & 0.567  & - &1.011 & freq & \\ \hline
    \multirow{4}{*}{office\_0} & \textbf{0.982} & 39.493 & 0.975 & 0.030 & \textbf{0.992} & 2.583 & hash & \multirow{4}{*}{Small} \\
     & 0.849 & 28.452 & 0.835 & 0.438 & 0.914 & 1.500 & freq & \\
     & - & \textbf{39.958} & \textbf{0.979} & \textbf{0.022}  & - &1.557 & hash & \\ 
     & - & 27.641 & 0.822 & 0.485  & - & \textbf{1.131} & freq & \\ \hline

    \multirow{2}{*}{room\_1} & - & \textbf{36.369} & \textbf{0.956} & \textbf{0.042} & - & \textbf{0.753} & hash & Small \\
    & - & 33.179 & 0.896 & 0.164 & - & 10.300 & freq & Large \\ \hline
    \end{tabular}
    \caption{
    Influence of positional encoding and model size on the quality of reconstruction of 3D scenes from Replica dataset.}
    \label{tab:table_encoding}
\end{table}

\subsection{Study of Different Uncertainties Usage in uSF Architecture}

We have compared the recovery quality of neural radiance field and neural semantic field in cases where the model predicts only color and semantic labels, as well as when the model additionally estimates only RGB uncertainty, only semantic uncertainty and both types of uncertainty. The metric values are given in Table \ref{dif_uncert_exps}. The experiments have been carried out on different train datasets sizes: small with 27 images and large with 562 images.

\begin{table}[t!]
    \centering
    \begin{tabular}{ c c c c c c }\hline
         \makecell{Training \\ sample size} & Metrics & \makecell{uSF \\ (w/o uncert.)} & \makecell{uSF \\ (rgb uncert.)} & \makecell{uSF \\ (semantic uncert.)} & \makecell{uSF \\ (both uncert.)} \\ \hline 
        \multirow{2}{*}{Small} & mIoU & 0.587 & 0.549 & \textbf{0.604} & 0.598\\
        & PSNR & 22.171 & 22.192 & \textbf{22.989} & 22.931 \\ 

        \multirow{2}{*}{Large} & mIoU & 0.975 & 0.975 & \textbf{0.977} & \textbf{0.977}\\
        & PSNR & 34.757 & 34.760 & \textbf{34.979} & 34.936 \\ \hline
    \end{tabular}
    \caption{Average values of mIoU and PSNR metrics for different numbers of images in the training dataset in cases where the model doesn't quantify uncertainty, estimates only rgb uncertainty, only semantic uncertainty, and both types of uncertainties.}
    \label{dif_uncert_exps}
    \medskip 
\end{table}

As we can see, in the case where there are 27 images in the training set the average PSNR value increases slightly with the addition of the uncertainty estimation. The average mIoU value improves in the case of predicting only semantic uncertainty and both uncertainties. In the case of a large set, adding a prediction also has a positive effect on metric values. So adding an uncertainty estimation does not degrade the quality of the network's predictions, and in some cases even improves it.

\begin{figure}[t!]
    \includegraphics[width=\linewidth]{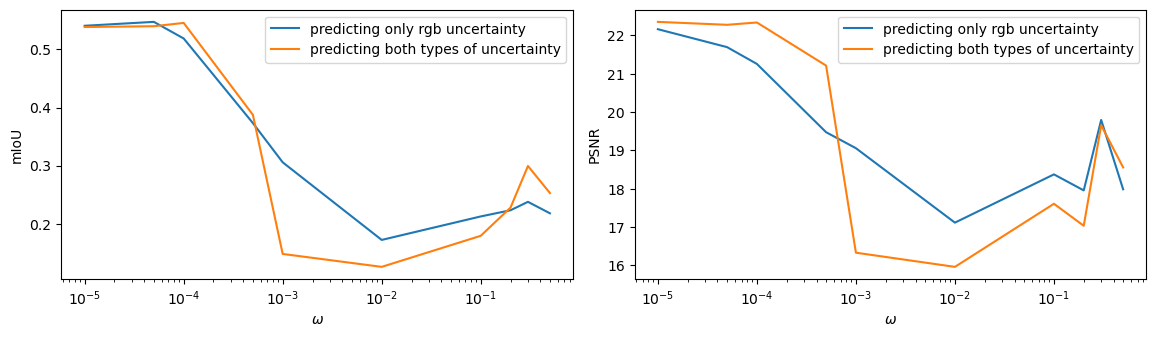}
    \includegraphics[width=\linewidth]{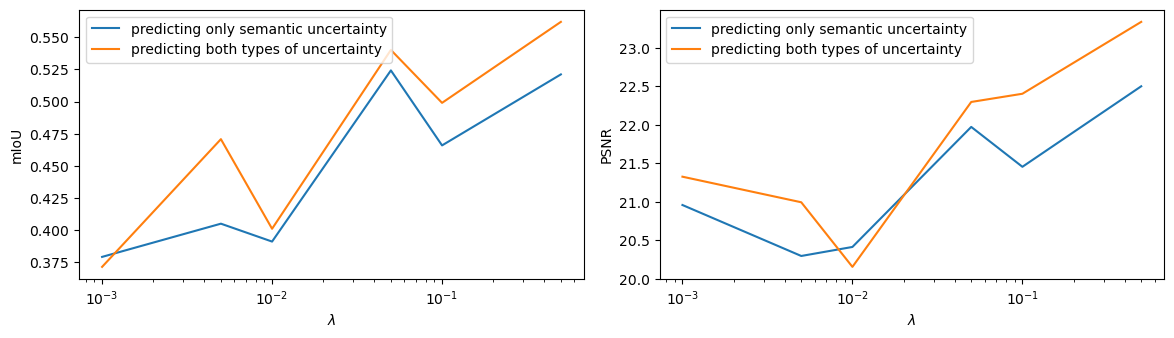}
    \caption{The mean values of mIoU and PSNR depending on $\omega$ in the first row and $\lambda$ in the second row for selected scenes from Replica Dataset.}
    \label{omega_lambda_plots}
\end{figure}

\subsection{Influence of the $\omega$ Hyperparameter on Semantic Field Reconstruction}

We have performed experiments to see how the quality of recovery depends on the parameter $\omega$. We fix $\lambda = 0.05$ and begin to change $\omega$ and carry out experiments on two different scenes that capture ordinary room and office environment. The small set of training images is used. At first, we predict only rgb uncertainty, after which we add prediction of semantic uncertainty.

The results are shown in the Fig.~\ref{omega_lambda_plots}.

As we can see with increasing of the $\omega$ parameter the quality of both color and semantic labels reconstruction is decreasing. Also the dependencies of metrics on the $\omega$ parameters have two local maxima where the first is $\omega \approx 10^{-5}$, and the second is $\omega \approx 0.3$. In the first case metrics are better, so in the following our experiments we consider $\omega$ close to $10^{-5}$.

\subsection{Influence of the $\lambda$ Hyperparameter on Semantic Field Reconstruction}

A similar experiment with fixed $\omega$ has been performed for the parameter $\lambda$. The results are shown in the Fig. ~\ref{omega_lambda_plots}. As we can see with increasing of the $\lambda$ parameter the quality of both color and semantic labels reconstruction is also increasing.

\subsection{Quality of the Semantic 3D-Scene Reconstruction}

We have compared the neural semantic fields reconstruction quality of our uSF method trained with the large (562) set of images with the Semantic NeRF method which is similar to our network without uncertainty estimations on the \textit{room\_1} scene. Our method have showed the best metric values for both predicting semantic labels and predicting color.

\begin{table}[t]
    \centering
    \begin{tabular}{ c c c c c }\hline
        Network & mIoU & LPIPS & Avg Acc & Total Acc \\ \hline
        Semantic NeRF & \,0.9313\, & \,0.2119\, & 0.9526 & 0.9903 \\
        uSF (both uncertainties) & 0.955 & 0.0757 & 0.975 & 0.996 \\ 
        uSF (rgb uncertainty) & 0.958 & \textbf{0.0726} & 0.9727 & \textbf{0.9977} \\ 
        uSF (semantic uncertainty) & \textbf{0.959} & 0.0731 & \textbf{0.976} & \textbf{0.9977} \\ 
         \hline
    \end{tabular}
    \caption{Quality of the semantic 3D-scene reconstruction of uSF and Semantic NeRF methods.}
    \label{comparison}
    \medskip
\end{table}

\section{Conclusion}

In this paper, we have presented a method for reconstructing the neural semantic field along with neural radiance field, which also provides the corresponding uncertainty for both color and semantic labels predictions. Under conditions of a limited number of images in the training set, the model estimating uncertainty shows better reconstruction results compared to the standard model without confidence associated with the predictions. It has been possible to increase the speed of the method by using a special positional encoding method. 

As directions for the development of our method, we see the implementation of epistemic uncertainty quantification for semantic labels prediction. We also want to develop the idea of using active learning for segmentation tasks.


\section*{FUNDING}

This work was supported by Russian Science Foundation, grant No. 20-71-10116, \url{https://rscf.ru/en/project/20-71-10116/.}

\section*{ETHICS APPROVAL AND CONSENT TO PARTICIPATE}

This work does not contain any studies involving human and animal subjects

\section*{CONFLICT OF INTEREST}

The authors of this work declare that they have no conflicts of interest.

\section*{References}

\small
\bibliographystyle{unsrt}
\renewcommand{\refname}{}
\nocite{*}
\bibliography{paper}
\vspace{-10mm}


\renewcommand{\refname}{СПИСОК ИСПОЛЬЗОВАННЫХ ИСТОЧНИКОВ}

\end{document}